\documentclass[lettersize,journal]{IEEEtran}
\usepackage{amsmath,amsfonts}
\usepackage{algorithmic}
\usepackage{algorithm}
\usepackage{array}
\usepackage[caption=false,font=normalsize,labelfont=sf,textfont=sf]{subfig}
\usepackage{textcomp}
\usepackage{stfloats}
\usepackage{url}
\usepackage{verbatim}
\usepackage{graphicx}
\usepackage{cite}
\usepackage{times}
\usepackage{soul}
\usepackage{url}
\usepackage[hidelinks]{hyperref}
\usepackage[utf8]{inputenc}
\usepackage[small]{caption}
\usepackage{amsthm}
\usepackage{booktabs}
\usepackage[switch]{lineno}
\usepackage{multirow}
\usepackage{amssymb}
\newtheorem{definition}{Definition}

\usepackage{colortbl}
\usepackage{xcolor} 
\definecolor{Gray}{rgb}{0.93,0.93,0.93}
\newcolumntype{g}{>{\columncolor{Gray}}c}
\usepackage[normalem]{ulem}
\useunder{\uline}{\ul}{}
\newcommand{\code}[0]{\url{https://github.com/ZzoomD/SaGIF}}

\begin{document}

\title{SaGIF: Improving Individual Fairness in Graph Neural Networks via Similarity Encoding}

\author{Yuchang Zhu, Jintang Li, Huizhe Zhang, Liang Chen, and Zibin Zheng,~\IEEEmembership{Fellow,~IEEE,}
\thanks{The research is supported by the National Key R\&D Program of China under grant No. 2022YFF0902500, the Guangdong Basic and Applied Basic Research Foundation, China (No. 2023A1515011050), Shenzhen Science and Technology Program (KJZD20231023094501003), and Tencent AI Lab (RBFR2024004). ({\itshape Corresponding author: Liang Chen.})}
\thanks{Liang Chen is with the School of Computer Science and Engineering, Sun Yat-Sen University, Guangzhou, China, 510007. Email: chenliang6@mail.sysu.edu.cn} 
\thanks{Yuchang Zhu and Huizhe Zhang are with the School of Computer Science and Engineering, Sun Yat-sen University, Guangzhou 510007, China.}
\thanks{Zibin Zheng and Jintang Li are with the School of Software Engineering, Sun Yat-sen University, Zhuhai  519082, China.}}

\markboth{Journal of \LaTeX\ Class Files,~Vol.~14, No.~8, August~2021}%
{Shell \MakeLowercase{\textit{et al.}}: A Sample Article Using IEEEtran.cls for IEEE Journals}

\IEEEpubid{0000--0000/00\$00.00~\copyright~2021 IEEE}

\maketitle

\begin{abstract}
    Individual fairness (IF) in graph neural networks (GNNs), which emphasizes the need for similar individuals should receive similar outcomes from GNNs, has been a critical issue. Despite its importance, research in this area has been largely unexplored in terms of (1) a clear understanding of what induces individual unfairness in GNNs and (2) a comprehensive consideration of identifying similar individuals. To bridge these gaps, we conduct a preliminary analysis to explore the underlying reason for individual unfairness and observe correlations between IF and \textit{similarity consistency}, a concept introduced to evaluate the discrepancy in identifying similar individuals based on graph structure versus node features. Inspired by our observations, we introduce two metrics to assess individual similarity from two distinct perspectives: topology fusion and feature fusion. Building upon these metrics, we propose \underline{S}imilarity-\underline{a}ware \underline{G}NNs for \underline{I}ndividual \underline{F}airness, named \textbf{SaGIF}. The key insight behind SaGIF is the integration of individual similarities by independently learning similarity representations, leading to an improvement of IF in GNNs. Our experiments on several real-world datasets validate the effectiveness of our proposed metrics and SaGIF. Specifically, SaGIF consistently outperforms state-of-the-art IF methods while maintaining utility performance. Code is available at: \code. 
\end{abstract}


\section{Introduction}
Graph neural networks (GNNs)~\cite{gcn,graphsage} have demonstrated remarkable performance, leading to their widespread application, e.g., recruitment~\cite{yang2022modeling}, recommendation systems~\cite{fan2019graph}, and drug discovery~\cite{li2021effective}. Despite their significant success, recent studies~\cite{fairgnn,nifty,edits} have revealed that GNNs face fairness issues due to biases inherent in the training data. Broadly, fairness issues in GNNs can be categorized into two types: group fairness~\cite{hardt2016equality} and individual fairness~\cite{dwork2012fairness}. Group fairness emphasizes the need for GNNs to make consistent predictions across different demographic groups divided by the sensitive attribute, e.g., age, and gender. In contrast, individual fairness (IF) focuses on a more fine-grained notion of fairness, aiming to ensure {\itshape similar predictions for similar individuals}. While IF is as crucial as group fairness in addressing fairness issues in GNNs, the exploration of IF is still in its infancy.

Dwork et al.~\cite{dwork2012fairness} first proposed the definition of IF: {\itshape similar individuals should be treated similarly}, and employed the Lipschitz condition to constrain the distance between similar individuals in the input and output spaces. However, little attention has been paid to improving individual fairness in graph-structured data. Generally, the mathematical foundation of existing IF methods~\cite{PFR,inform,redress,guide,gfairhint} for graph-structured data is based on the Lipschitz condition. Given a pair of individuals ($u$, $v$), the Lipschitz condition stipulates that the output distance $D_{2}(u, v)$ is upper bounded by the product of the input distance $D_{1}(u, v)$ and a constant $L$. Here, $D_{1}(\cdot)$ and $D_{2}(\cdot)$ are the distance metric in the input space and the output space, respectively, while $L$ is known as the Lipschitz constant. Following the Lipschitz condition, these studies aim to construct distance measurements on output space from various perspectives, including Euclidean distance~\cite{PFR}, a ranking perspective~\cite{redress}, and IF with group consideration~\cite{guide}. Based on these distance measurements in the output space, the core idea behind these methods is to minimize the inconsistency of pairs of individuals between the input and output spaces. 

\IEEEpubidadjcol
Despite the advancements in IF for graph-structured data, two critical aspects remain under-explored in existing methods. Firstly, these methods primarily focus on defining the distance (or similarity) measurement in the output space, while neglecting the definition in the input space. The similarity measurement in the input space uses the input graph to compute a pairwise similarity matrix, known as the oracle similarity matrix, which is essential for identifying similar nodes. Existing methods typically construct the oracle similarity matrix by calculating cosine similarity from features or using the Jaccard index with the adjacency matrix~\cite{inform,redress}. However, due to the complex nature of graph-structured data, these approaches often rely solely on node features or graph topology, leading to inaccurate identification of similar nodes. For example, in a university social network where nodes represent students and edges represent friendships, two classmates might have similar friendship patterns if only graph topology is considered. However, their node features could be very different due to varying backgrounds such as family and region. Therefore, relying solely on node features or graph topology results in inaccurate identification of similar nodes. Secondly, most current research simply adopts widely used fairness techniques that follow the Lipschitz condition framework, without a deep understanding of the underlying causes of individual unfairness in graph-structured data. 

To address the issues mentioned above, we first conduct a preliminary study to explore the underlying reasons behind individual unfairness. Experimental results suggest that individual unfairness primarily stems from a similarity inconsistency between node features and graph topology. Specifically, the lists of similar nodes for a given node differ when similarity is defined based solely on features versus graph topology. Motivated by this observation, we propose two measurement metrics to calculate the oracle similarity matrix, considering both node features and graph topology. Building upon our proposed metrics, we further propose a \underline{S}imilarity-\underline{a}ware \underline{G}NN for \underline{I}ndividual \underline{F}airness, namely, \textbf{SaGIF}. Benefiting from the design of similarity encoding, SaGIF is compatible with various GNN architectures and incurs only a linear complexity cost. Thus, our contributions can be summarized as follows:
\begin{itemize}
    \item We study the reasons for individual unfairness in graphs and propose two metrics to calculate the oracle similarity matrix considering both node features and topology.
    \item We introduce an IF framework for learning fair GNNs, named SaGIF. The key insight of SaGIF is the preservation of oracle similarity through similarity encoding.
    \item We conduct extensive experiments to validate our proposed method. The results demonstrate that SaGIF improves the IF level in GNNs while preserving their utility.
\end{itemize} 

\section{Related Work}
\subsection{Fairness in Machine Learning}
In machine learning, fairness issues highlight that decision-making by machine learning models is free from discrimination or favoritism towards individuals or demographic groups based on their sensitive attributes~\cite{mehrabi2021survey}. While various fairness notions have been proposed in machine learning, these can generally be summarized into three types: {\itshape group fairness}~\cite{hardt2016equality}, {\itshape other fairness}~\cite{kusner2017counterfactual}, and {\itshape individual fairness}~\cite{dwork2012fairness}. In this work, we focus on individual fairness.

{\itshape Group fairness} is the most widespread notion of fairness, which encourages equally treating different demographic groups defined by the sensitive attribute~\cite{hardt2016equality}. To improve group fairness, existing technologies encompass areas such as adversarial learning~\cite{madras2018learning,fairgnn}, fairness regularization~\cite{zeng2021fair}, and distribution alignment~\cite{li2023fairer}. {\itshape Other fairness} encompasses a variety of definitions, each addressing specific aspects of fairness. For instance, Accuracy-parity-style fairness~\cite{ma2021subgroup} emphasizes the accuracy disparity across different subgroups, which also can be summarized as a form of group fairness. Max-Min fairness~\cite{rawls2001justice,hashimoto2018fairness} focuses on improving per-group performance. Counterfactual fairness~\cite{ma2023learning} aims to guarantee consistent predictions for each individual and its counterfactuals. Degree-related fairness~\cite{kang2022rawlsgcn} seeks to balance the utility between high-degree and low-degree nodes.

{\itshape Individual fairness}, first defined in Dwork et al.'s work~\cite{dwork2012fairness}, advocates the principle that {\itshape similar individuals should be treated similarly}. Additionally, Dwork et al.~\cite{dwork2012fairness} employ the Lipschitz condition to constrain similar individuals' distance between input and output space. In line with this condition, PFR~\cite{PFR} utilizes side information about equally deserving individuals to construct a fairness graph and then learns pairwise fair representations. Another research line aims to balance the level of individual fairness with the utility of the downstream task. For example, LFR~\cite{zemel2013learning} formulates fairness as an optimization problem, combining objectives to achieve a trade-off between utility and fairness performance. iFair~\cite{lahoti2019ifair} probabilistically maps user records into a low-rank representation to reconcile individual fairness with application utility. Li et al.~\cite{li2023accurate} introduce a siamese fairness approach to optimize what they call {\itshape accurate fairness}, aligning individual fairness with accuracy. Moreover, some studies~\cite{mukherjee2020two,john2020verifying} verify the level of the individual fairness of models, filling up the absence of individual fairness and its bias measurement metrics. However, these prior efforts are tailored for structured data and may not be suitable for graph-structured data due to the inherent difference between the two. Different from these efforts, our work delves into the issues of individual fairness in graph-structured data and GNNs.

\subsection{Individual Fairness in Graphs}
Owing to the ubiquity of graphs, there has been a growing focus on addressing fairness issues in graphs. Despite its critical importance, individual fairness in graphs has not attracted as much attention as group fairness. In this context, InFoRM~\cite{inform} emerges as the pioneering work in addressing the IF issue in the realm of graph mining, establishing a fairness framework anchored in the Lipschitz property. GUIDE~\cite{guide} observes a gap in the current literature that has significant disparities in IF levels among different groups. Utilizing the same mathematical foundation as InFoRM, GUIDE advances IF by reducing group-level disparity through learnable personalized attention. Different from these approaches, REDRESS~\cite{redress} investigates the IF issue from a ranking perspective, striving to ensure consistency in ranking lists for individuals across input and output spaces. Meanwhile, GFairHint~\cite{gfairhint} encodes the similarity through an auxiliary link prediction task and proposes a universal IF framework. 
However, there remains a drawback in existing research regarding a graph-tailored metric for individual similarity and a thorough understanding of the source of individual unfairness in graphs. Unlike previous studies that overlook comprehensive individual similarity, our work measures it from both topology and node features, as shown in Section~\ref{sec:sim_metric}. SaGIF ensures individual fairness through learnable positional encoding, termed similarity encoding, offering a novel perspective with linear complexity. 

\section{Preliminary Analysis}
\subsection{Notations}
Let $\mathcal{G}=(\mathcal{V}, \mathcal{E}, \textbf{X})$ denote an attributed and unweighted graph, where $\mathcal{V}$ is a set of $\lvert \mathcal{V} \rvert = n$ nodes, $\mathcal{E}$ is a set of $\lvert \mathcal{E} \rvert = m$ edges, $\textbf{X} \in \mathbb{R}^{n\times d}$ is the node attribute matrix. $\textbf{A} \in \{0,1\}^{n\times n}$ is the adjacency matrix. The diagonal matrix $\textbf{D} \in \mathbb{R}^{n\times n}$ is the degree matrix of $\textbf{A}$. $\textbf{Y}$ and $\hat{\textbf{Y}}$ denote the ground truth and prediction for a specific downstream task, respectively. $\textbf{S} \in \mathbb{R}^{n\times n}$ and $\hat{\textbf{S}} \in \mathbb{R}^{n\times n}$ denote the oracle similarity matrix and the similarity matrix in the output space, where $\textbf{S}_{ij}$ is the similarity between node $v_{i}$ and $v_{j}$. 

In the field of IF in graphs, prior efforts~\cite{guide,petersen2021post} utilize the Lipschitz condition as the mathematical foundation to develop fair models. The Lipschitz condition is given as follows:
\begin{definition}
    \textbf{Lipschitz condition}. A function $f(\cdot)$ satisfies the Lipschitz condition if for any two individuals $x$, $y$, the following inequality holds
        \begin{equation}
            \label{eq:Lipschitz}
            D_{2}(f(x),f(y)) \leq LD_{1}(x,y), \forall(x,y),
        \end{equation}
    where $D_{1}$ and $D_{2}$ measure the distance between $x$, $y$ in the input space and the output space, respectively. The constant $L$ is known as the Lipschitz constant.
\end{definition}

Based on Eq.~\eqref{eq:Lipschitz}, the fairness optimization objective of existing approaches can be summarized as follows:
\begin{equation}
    \small
    \label{eq:loss_sota}
        \sum_{v_{i} \in \mathcal{V}}\sum_{v_{j} \in \mathcal{V}}{D_{2}(f(v_{i}),f(v_{j}))\mathbf{S}[i,j]}=2\text{Tr}(\hat{\mathbf{Y}}^{T}\mathbf{L}_{\mathbf{S}}\hat{\mathbf{Y}}) \leq m\epsilon,
\end{equation}
where $\hat{\textbf{Y}}$ is the outcome of models with $\hat{\textbf{Y}}[i,:]=f(v_{i})$ and $\hat{\textbf{Y}}[j,:]=f(v_{j})$. $\textbf{L}_{\textbf{S}}$ is the Laplacian matrix of the oracle similarity matrix \textbf{S}. $m$ is the number of non-zero elements in \textbf{S}. $\epsilon$ is a constant for tolerance. Eq.~\eqref{eq:loss_sota} serves as a relaxation of Eq.~\eqref{eq:Lipschitz}, embodying the principle that nodes similar in the input space should exhibit similar characteristics in the output space. Building upon Eq.~\eqref{eq:loss_sota}, existing approaches focus on defining $D_{2}$ from various perspectives. 

\begin{figure*}[!htb]
    \centering
    \includegraphics[width=0.9\textwidth]{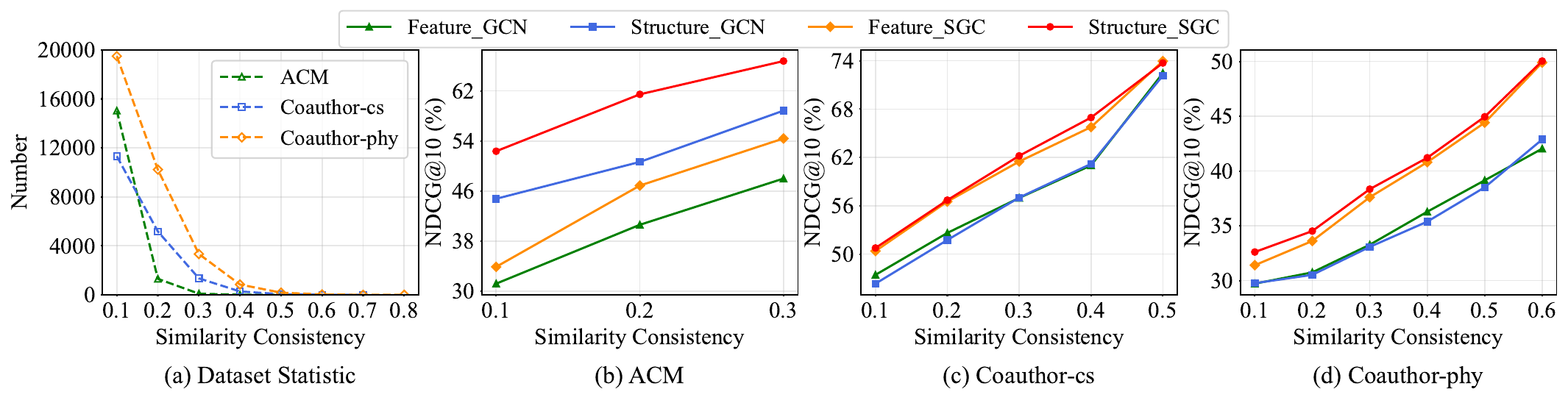}
    \caption{The preliminary analysis on three datasets. (a) Similarity consistency statistic of three datasets. (b)-(d) Node classification results, indicating the correlation between the similarity consistency and individual fairness (NDCG@10). A higher value of NDCG@10 indicates a higher level of IF. Different polylines are labeled as “$\textbf{S}$'s Calculation Perspective (Feature or Structure)$\_$backbone (GCN or SGC)”}
    \label{fig:preliminary}
\end{figure*}

\subsection{Source of Individual Unfairness}
\label{sec:pre}
While prior efforts have advanced the level of IF in graphs, the underlying causes of individual unfairness remain unknown. Considering the superior performance of GNNs, our investigation centers on the origins of individual unfairness in conjunction with the fundamental mechanism of most GNNs, i.e., the message-passing mechanism. Specifically, the message-passing mechanism updates the representation of the target node by aggregating messages from neighbors.

To facilitate our investigation, we introduce the concept of \textit{similarity consistency}, which fundamentally refers to the degree of discrepancy in identifying similar individuals through node features and graph topology, respectively. For a given node $v_{i}$, a higher similarity consistency suggests that the identification of nodes similar to $v_{i}$ remains uniform, whether based on node features or graph topology. Formally, we define similarity consistency as follows.
\begin{definition}
\label{def:sc}
    \textbf{Similarity consistency}. Given two oracle similarity matrices $\textbf{S}_{f}$, $\textbf{S}_{t}$ calculated from node features and graph topology, sorting the entries of $\textbf{S}_{f}[i,:]$ and $\textbf{S}_{t}[i,:]$ for node $v_{i}$ in descending order to acquire top-$k$ node index sets $\textbf{s}_{v_i}^{f}$ and $\textbf{s}_{v_i}^{t}$. Consequently, the similarity consistency of the node $v_{i}$ can be formulated as follows:
    \begin{equation}
        \label{eq:consistency}
        c_{v_{i}} = \frac{|\textbf{s}_{v_i}^{f} \cap \textbf{s}_{v_i}^{t}|}{k},
    \end{equation}
    where $k \ll n$ is an integer hyperparameter. Since $\textbf{s}_{v_i}^{f}$ and $\textbf{s}_{v_i}^{t}$ include top-$k$ positions exhibiting the highest similarity to $v_{i}$, similarity consistency measures the disparity in identifying similar nodes based on node features as compared to graph topology for a given node.
\end{definition}

Based on Definition~\ref{def:sc}, we analyze similarity consistency across three datasets (ACM~\cite{tang2008arnetminer}, coauthor-cs, and coauthor-phy~\cite{shchur2018pitfalls}). According to Eq.~\eqref{eq:consistency}, we calculate the pairwise cosine similarities of the node attribute matrix and the adjacency matrix to obtain oracle similarity matrices $\textbf{S}_{f}$, $\textbf{S}_{t}$, respectively. We set $k$ to 10 and count the number of nodes corresponding to each similarity consistency level, ranging from 0.1 to 1.0. As shown in Figure~\ref{fig:preliminary}(a), the majority of nodes in these datasets exhibit relatively low similarity consistency, indicating significant discrepancies between identifying similar nodes from the perspectives of node features and topology. 

To further investigate the correlation between similarity consistency and individual unfairness, we conduct the node classification experiment. We set $k$ to 10 and utilize two commonly used GNNs as backbones, i.e., GCN~\cite{gcn} and SGC~\cite{sgc}. In line with prior efforts~\cite{redress,gfairhint}, we utilize a ranking metric NDCG@$k$~\cite{ndcg} to evaluate the IF level in model outcomes among nodes with varying degrees of similarity consistency. NDCG@$k$ measures the similarity between the ranking list from the oracle similarity matrix \textbf{S} and the similarity matrix $\hat{\textbf{S}}$ in the output space. We take the node feature matrix \textbf{X} and the adjacency matrix \textbf{A} as input to calculate \textbf{S}, respectively, referred to as feature and structure perspectives. To avoid the impact of randomness, we only evaluate the IF levels of the similarity consistency groups with more than 10 nodes. In Figure~\ref{fig:preliminary}(b)-(d), different polylines are labeled as ``$\textbf{S}$'s Calculation Perspective (Feature or Structure)$\_$backbone (GCN or SGC)". For instance, ``Feature$\_$GCN" represents the calculation of \textbf{S} from the feature perspective, employing GCN as the backbone. As shown in Figure~\ref{fig:preliminary}(b)-(d), all experiments demonstrate the correlation between the similarity consistency and IF. Notably, the level of IF in model predictions exhibits an incremental trend in correlation with the rise in similarity consistency. In instances where nodes exhibit high similarity consistency, the predictive behavior of the model more closely aligns with the principle that similar nodes receive similar treatment. 

We utilize the message-passing mechanism to interpret our observation. GNNs aggregate messages (node features) according to the graph structure, empirically proving that the representation of connected nodes may be more similar after aggregations~\cite{chiang2019cluster,wang2020unifying}. In cases of low similarity consistency, where there is a big difference in identifying similar nodes based on features versus topology, nodes sharing similar features might be spatially distant, while being proximate to their immediate neighbors in the graph. This principle similarly applies to nodes with a similar graph topology (or 1-hop neighbors). The model's predictions increasingly diverge from the ideal of IF. In summary, the source of individual unfairness in GNNs can be traced back to the issue of similarity inconsistency. Owing to the disparities in similarity calculations derived from diverse perspectives, there arises a demand for a comprehensive similarity metric that incorporates multiple factors for a more accurate ``similar individuals" assessment.

Additionally, in the ACM dataset, the NDCG@k values derived from the structural perspective (red and blue lines) are closer, while in the Coauthor-CS and Coauthor-Phy datasets, the NDCG@k values using SGC as the backbone (red and orange lines) are more similar. This can be explained as follows: The ACM dataset, a citation network with a strong community pattern, relies heavily on structural information, resulting in similar NDCG@k values. In contrast, the Coauthor datasets have complex topologies that shallow GNNs struggle to capture, making node features more influential and leading to similar performance within the same backbone.

\section{Similarity Metrics}
\label{sec:sim_metric}
\begin{figure}[!t]
    \centering
    \includegraphics[width=\linewidth]{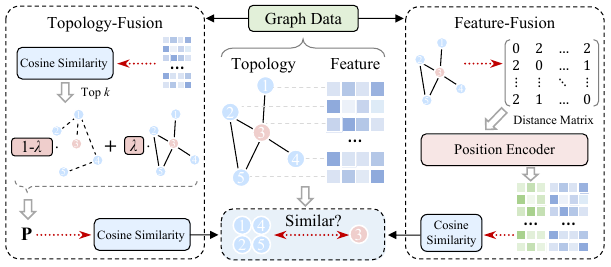}
    \caption{The illustration of similarity measurement from two perspectives (topology fusion and feature fusion).}
    \label{fig:sim_measure}
\end{figure}

Previous studies~\cite{inform,redress,guide} calculate the oracle similarity matrix solely based on either node features or graph structure, which often yields an impractical list of similar individuals. Relying on such a list of similar individuals, any methodologies would deviate significantly from achieving IF. Since real-world graph-structured data often forms attributed networks~\cite{pfeiffer2014attributed}, comprising both node features and edges, it is appropriate to integrate node features and edge information. In light of the analysis in Section~\ref{sec:pre}, we introduce two similarity metrics, namely, \textit{topology fusion} and \textit{feature fusion}, to bridge this gap. These two metrics are shown in Figure~\ref{fig:sim_measure}.

\begin{figure*}[!th]
    \centering
    \includegraphics[width=0.8\textwidth]{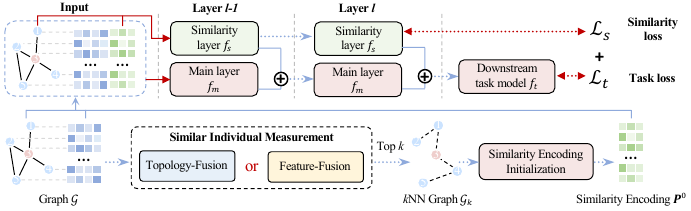}
    \caption{The overview of our proposed method SaGIF.}
    \label{fig:architecture}
\end{figure*}

\subsection{Topology Fusion}
\label{subsec:t_f}
Inspired by the node similarity preserving aggregation~\cite{jin2021node}, topology fusion aims to integrate the similarity information of node features and graph structure into a propagation matrix \textbf{P}. Based on \textbf{P}, a natural way for constructing \textbf{S} is to calculate row-wise similarity through cosine similarity. Given a graph $\mathcal{G}=(\mathcal{V}, \mathcal{E}, \textbf{X})$ with an adjacency matrix \textbf{A} and a degree matrix \textbf{D}, the node features matrix \textbf{X} can be transformed into a $k$NN graph through the following process: (1) compute pairwise similarity for each node pair $(v_{i}, v_{j})$ via cosine similarity. (2) For each node $v_i$, select the top $k$ nodes with the highest similarity to form connections with $v_i$. Based on the $k$NN graph, the adjacency matrix $\textbf{A}_{k}$ and its degree matrix $\textbf{D}_{k}$ are derived. Thus, \textbf{P} can be formulated as follows:
\begin{equation}
    \label{eq:prop_mat}
    \textbf{P} = \underbrace{\lambda\tilde{\textbf{D}}^{-1/2}\tilde{\textbf{A}}\tilde{\textbf{D}}^{-1/2}}_{Structure} + \underbrace{(1-\lambda)\textbf{D}_{k}^{-1/2}\textbf{A}_{k}\textbf{D}_{k}^{-1/2}}_{Feature},
\end{equation}
where $\tilde{\textbf{A}}=\textbf{A}+\textbf{I}$ and $\tilde{\textbf{D}}$ is the degree matrix of $\tilde{\textbf{A}}$. $\lambda$ is a hyperparameter to balance the similarity information between the original graph and the $k$NN graph (node features). According to \textbf{P}, the ($i$, $j$)-th entry of the oracle similarity matrix $\textbf{S} \in \mathbb{R}^{|\mathcal{V}|\times|\mathcal{V}|}$ can be calculated as follows:
\begin{equation}
    \label{eq:ora_mat_cal}
    \textbf{S}_{ij} = \frac{\textbf{P}[i, :]^{T}\textbf{P}[j, :]}{\|\textbf{P}[i, :]\|\|\textbf{P}[j, :]\|}.
\end{equation}
\subsection{Feature Fusion}
In contrast to topology fusion, feature fusion incorporates all similarity information into a single synthetic features matrix. This process is particularly relevant in the context of recent advancements in position-aware GNNs~\cite{liu2023position,you2019position}. Specifically, feature fusion converts the graph structure into a positional embedding using a position encoder. It then constructs a synthetic features matrix by concatenating this positional embedding with the node features matrix. Utilizing this matrix, \textbf{S} is calculated as detailed in Eq.~\eqref{eq:ora_mat_cal}. Given a graph $\mathcal{G}=(\mathcal{V}, \mathcal{E}, \textbf{X})$, the distance matrix $\textbf{M} \in \mathbb{R}^{|\mathcal{V}|\times|\mathcal{V}|}$ is derived using Dijkstra’s algorithm~\cite{dijkstra2022note}. Here, $\textbf{M}_{ij}$ represents the length of the shortest path between node $v_i$ and $v_j$. Additionally, the diagonal elements of \textbf{M} are set to 0, and elements corresponding to unconnected node-pairs are set to $\infty$. Taking \textbf{M} as input, the position encoder generates the positional embedding in two steps: (1) value mapping and (2) principal component analysis (PCA) transformation. In value mapping, following the setting Liu's work~\cite{liu2023position}, we map the values of the distance matrix to a small range to mitigate the distance variance. Formally, the mapping function can be formulated as follows:
\begin{equation}
    \label{eq:cos_map}
        \tilde{\textbf{M}}_{ij}=\left\{
        \begin{aligned}
            & \cos(\frac{\pi}{\max(\textbf{M}[i, :])}\times\textbf{M}_{ij}) \quad &,\textbf{M}_{ij}\neq\infty \\
            &-1.5 \quad &,\textbf{M}_{ij}=\infty,
        \end{aligned}
        \right.
\end{equation}
where $\max(\textbf{M}[i, :])$ is the maximum of the $i$-th row of \textbf{M}, excluding any $\infty$ elements. This value indicates the maximum length of the shortest path from node $v_i$ to all reachable nodes. According to Eq.~\eqref{eq:cos_map}, values on a large scale can be mapped to the range [-1, 1]. Specifically, a relatively small distance, e.g., 1, will be mapped to a value as close to 1 as possible. A large distance, e.g., $\max(\textbf{M}[i, :])$, will be mapped to -1. 

In the PCA transformation, the dimensionality of $\tilde{\textbf{M}}_{ij}$ is reduced to $|\mathcal{V}|\times d_{sim}$ through PCA. Denoting the positional embedding obtained after PCA as $\overline{\textbf{M}}$ with a dimensionality of $d_{sim}$, the PCA process can be formulated as follows:
\begin{equation}
    \label{eq:pca}
        \overline{\textbf{M}} = \operatorname{PCA}(\tilde{\textbf{M}}).  
\end{equation}

By concatenating $\overline{\textbf{M}}$ with node feature matrix $\textbf{X}$, a synthetic feature matrix is obtained. Consequently, $\textbf{S}$ can be calculated as defined in Eq.~\eqref{eq:ora_mat_cal}. Overall, topology fusion incorporates all similarity information into a propagation matrix and calculates $\textbf{S}$ in the topology aspect. Conversely, feature fusion encodes similarity information from graph structure as position information while calculating $\textbf{S}$ with the synthetic features matrix.  

\section{Present Work}
Based on our proposed metrics in Section~\ref{sec:sim_metric}, ``similar individual'' can be accurately identified. Our observation in Section~\ref{sec:pre} suggests that a straightforward approach to improve IF involves improving similarity consistency by modifying node features or graph structure. However, it inevitably leads to a deterioration in utility performance. Meanwhile, prior efforts also suffer from the sacrifice of utility due to the enforcement of fairness in learning processes. Drawing insights from position-aware GNNs research~\cite{dwivedi2021graph}, we propose SaGIF, specifically designed to improve individual fairness while overcoming the sacrifice in utility. Similar to the learning of positional representations, the core idea behind SaGIF is to employ independent similarity learning to preserve similarity information as measured by our proposed metrics. The overview of SaGIF is detailed in Figure~\ref{fig:architecture}.

\subsection{Framework}
Following the framework of message-passing GNNs, SaGIF $f$ consists of a downstream task model and two independent encoders, named the main encoder and the similarity encoder. Given an attributed graph $\mathcal{G}=(\mathcal{V}, \mathcal{E}, \textbf{X})$, for a given node $v_i$, the update equation in the $l$-th layer of a standard message-passing-based GNN can be summarized as follows:
\begin{equation}
    \label{eq:mpgnn}
    \textbf{h}^{l}_{v_i} = f_{g}(\textbf{h}^{l-1}_{v_i}, \{\textbf{h}^{l-1}_{v_j}\}_{v_j\in\mathcal{N}(v_i)}),    
\end{equation}
where $f_{g}$ represents a graph convolutional function with learnable parameters, varying with different GNNs' designs. $\textbf{h}^{l}_{v_i}, \textbf{h}^{l-1}_{v_i}, \textbf{h}^{l-1}_{v_j} \in \mathbb{R}^{d^{'}}$ are node representations in the $l$-th and $l$-1-th layer, respectively. $d^{'}$ represents the hidden dimension. $\mathcal{N}(v_i)$ represents the set of nodes adjacent to node $v_i$ in $\mathcal{G}$.

To improve IF, SaGIF preserves similarity information among ``similar individuals'' using an idea analogous to positional encoding. This idea is inspired by a recent study~\cite{dwivedi2021graph} advocating for the decoupling of structural and positional representations to yield more expressive node representations. In our scenarios, SaGIF achieves the decoupling of structural and similarity representations through the learning of independent similarity representations. Specifically, based on Eq.~\eqref{eq:mpgnn}, the $l$-th layer of SaGIF can be formulated as follows:
\begin{align}
    \begin{split}
        \label{eq:sagif}
        \textbf{h}^{l}_{v_i} &= f_{m}([\textbf{h}^{l-1}_{v_i}, \textbf{p}^{l-1}_{v_i}], \{[\textbf{h}^{l-1}_{v_j},\textbf{p}^{l-1}_{v_j}]\}_{v_j\in\mathcal{N}(v_i)}), \\
        \textbf{p}^{l}_{v_i} &= f_{s}(\textbf{p}^{l-1}_{v_i}, \{\textbf{p}^{l-1}_{v_j}\}_{v_j\in\mathcal{N}(v_i)}), 
    \end{split}
\end{align}
where $f_{m}$ and $f_{s}$, which denote the main layer and similarity layer respectively, are two graph convolutional functions that have identical structures but operate independently, similar to $f_{g}$. $\textbf{p}^{l}_{v_i}\in \mathbb{R}^{d_{sim}}$ is the similarity representation of node $v_i$ in the $l$-th layer. [$\cdot$] is the concatenation operation. 
When $l=0$, $\textbf{h}^{0}_{v_i}=\textbf{X}[i,:]$ and $\textbf{p}^{0}_{v_i}=\textbf{P}^{0}[i,:] \in \mathbb{R}^{d_{sim}}$ is the initial similarity encoding deriving from the oracle similarity metric \textbf{S} calculated through our proposed metrics, as detailed in the next section. Taking the concatenation of $\textbf{h}^{l}_{v_i}$ and $\textbf{p}^{l}_{v_i}$ as input, the downstream task model $f_t$, e.g., a GNN classifier or a linear layer, makes predictions for node $v_i$.

\subsection{Similarity Encoding Initialization}
\label{subsec:init_SE}
SaGIF follows the idea of MPGNNs-LSPE~\cite{dwivedi2021graph} to obtain the initial similarity encoding $\textbf{P}^{0}\in \mathbb{R}^{|\mathcal{V}|\times d_{sim}}$. The generation of initial similarity encoding can be divided into two stages: $k$NN graph construction and Laplacian similarity encoding. Given an attributed graph $\mathcal{G}=(\mathcal{V}, \mathcal{E}, \textbf{X})$, the oracle similarity matrix \textbf{S} can be calculated as described in Section~\ref{sec:sim_metric}. During the $k$NN graph construction, for each node $v_i$, based on \textbf{S}, we select the top $k$ nodes exhibiting the highest similarity with $v_i$ to form connections. Consequently, a $k$NN graph $\mathcal{G}_k=(\mathcal{V}, \mathcal{E}_k)$ and its adjacency matrix $\textbf{A}_k$ are generated. Unlike the $k$NN graph employed in Eq.~\eqref{eq:prop_mat}, this $k$NN graph is uniquely characterized by both node features and graph structure, thereby encoding the similarity information in $\mathcal{G}$ as graph structural information. In the stage of Laplacian similarity encoding, Laplacian eigenvector, a commonly used technique for graph positional encoding, is employed to obtain the initial similarity encoding due to its nature of distance sensitivity. This process entails the factorization of the Laplacian matrix $\textbf{L}_{k}$ of $\mathcal{G}_k$, and can be formulated as follows:
\begin{equation}
    \label{eq:sim_enc_init}
    \textbf{L}_{k} = \textbf{I}-\textbf{D}_{k}^{-1/2}\textbf{A}_{k}\textbf{D}_{k}^{-1/2}=\textbf{U}^{T}\Lambda\textbf{U}
\end{equation}
where $\textbf{D}_{k}\in \mathbb{R}^{|\mathcal{V}|\times|\mathcal{V}|}$ is the degree matrix of $\textbf{A}_{k}$, $\textbf{I}\in \mathbb{R}^{|\mathcal{V}|\times|\mathcal{V}|}$ is a identity matrix. $\textbf{U},\Lambda \in \mathbb{R}^{|\mathcal{V}|\times|\mathcal{V}|}$ represent the eigenvectors and eigenvalues, respectively. Thus, the initial similarity encoding $\textbf{p}^{0}_{v_i}=\textbf{U}[i, :d_{sim}]\in \mathbb{R}^{d_{sim}}$.
\subsection{Optimization Objective and Algorithm}
Due to the coupling of the representation of nodes and its similarity representation, we can adapt the independent loss function for the downstream task and similarity encoding. To optimize the learning of similarity representations, we utilize a similarity-based loss to ensure the preservation of similarity information. Thus, the overall optimization objectives of SaGIF can be summarized as follows: 
\begin{equation}
\label{eq:loss}
    \min_{\theta_{f}} \mathcal{L}=\mathcal{L}_{t}+\alpha\mathcal{L}_{s},
\end{equation}
where $\theta_{f}=\{\theta_{f_{m}}, \theta_{f_{s}}, \theta_{f_{t}}\}$ is the parameter set of $f_{m}$, $f_{s}$, and $f_{t}$. $\mathcal{L}_{t}$ is the downstream task loss function and can be a cross-entropy function for the node classification task. $\alpha$ is a hyperparameter to balance utility and fairness. 
$\mathcal{L}_{s}$ is the similarity-based loss, which can be formulated as follows:
\begin{equation}
\label{eq:sim_loss}
    \mathcal{L}_{s}(\textbf{P}, \textbf{S})=\frac{1}{k|\mathcal{V}_{train}|}\|D(\textbf{P}, \textbf{P}^T)-\textbf{S}\|_{F}^{2},
\end{equation}
where $\textbf{P}=\{\textbf{p}_{v_{i}}^{l}\}_{v_{i}\in\mathcal{V}_{train}}$ is the similarity representations of training nodes $\mathcal{V}_{train}$. $D(\cdot)$ denotes a cosine similarity metric, calculating the similarity between each row within \textbf{P} with one another. $\|\cdot\|_{F}$ is the Frobenius norm. 

We present a summary of the SaGIF training pipeline for node classification in Algorithm~\ref{alg:training}. Apart from initializing similarity encoding according to Eq.~\eqref{eq:sim_enc_init}, an alternative approach involves employing random walks~\cite{mialon2021graphit}. A comparative analysis of these two initializations has been conducted in our experiments, with further details presented in Section~\ref{subsec:probe}. According to Algorithm~\ref{alg:training}, the similarity information preserved by SaGIF is based on the oracle similarity matrix \textbf{S}. As described in Section~\ref{sec:sim_metric}, topology fusion directly integrates the original graph topology with a similarity topology derived from node features. On the other hand, feature fusion extracts distance information from the original topology, and integrates it with node feature information. Thanks to such topology pattern-independent approaches, both methods are unaffected by whether the original graph topology exhibits homophily or heterophily. Consequently, this enhances the scalability of SaGIF across different types of datasets.


\begin{algorithm}[t]
\caption{Training Algorithm of SaGIF}
\label{alg:training}
    \begin{flushleft}
        \textbf{Input}: $\mathcal{G}=(\mathcal{V}, \mathcal{E}, \textbf{X})$, node label ground truth $\textbf{Y}$, the SaGIF model $f$, including the downstream task model, the main encoder, and the similarity encoder, hyperparameters $epoch$, $\alpha$, and $k$. \\
        \textbf{Output}: Learned SaGIF model $f$.
    \end{flushleft}
    \begin{algorithmic}[1] 
    \STATE // Initial Similarity Encoding
    \STATE Calculate the oracle similarity
matrix $\textbf{S}$ via {\itshape Topology Fusion} or {\itshape Feature Fusion} in Section~\ref{sec:sim_metric};
    \STATE Construct a $k$NN graph $\mathcal{G}_{k}$ according to $\textbf{S}$;
    \STATE Calculate initial similarity encoding $\textbf{P}^{0}$ via Eq.~\eqref{eq:sim_enc_init};

    \STATE // model training
    \FOR{$t=1$ to $epoch$}
        \STATE $\hat{\textbf{Y}} \leftarrow f(\mathcal{G},\textbf{P}^{0})$;
        \STATE Calculate $\mathcal{L}_{t}$ via $\hat{\textbf{Y}}$ and cross-entropy function;
        \STATE Calculate $\mathcal{L}_{s}$ via $\textbf{S}$ and Eq.\eqref{eq:sim_loss};
        \STATE Calculate loss function $\mathcal{L} \leftarrow \mathcal{L}_{t}+\alpha\mathcal{L}_{s}$;
        \STATE Update parameters of $f$ by gradient descent;  
    \ENDFOR
        \STATE \textbf{return} $f$;
    \end{algorithmic}
\end{algorithm}

\subsection{Complexity Analysis}
We briefly analyze the computational complexity of SaGIF. Assuming GCN serves as the graph convolutional function, a standard message-passing-based GNN with two layers exhibits a complexity of $\mathcal{O}(\lvert \mathcal{E} \rvert d{d^{'}}^{2})$ due to sparse-dense matrix multiplications, indicating linear complexity with respect to the number of graph edges. As shown in Eq.~\eqref{eq:sagif}, SaGIF shares a similar architecture with learnable structural and positional encodings of~\cite{dwivedi2021graph}, which also results in linear complexity.

\section{Experiments}
We conduct the node classification task experiment on six commonly used datasets, including Cora, Citeseer, Pubmed~\cite{yang2016revisiting}, ACM~\cite{tang2008arnetminer}, coauthor-cs, and coauthor-phy~\cite{shchur2018pitfalls}. The statistics of datasets are shown in Table~\ref{tab:statistic}. 
We compare SaGIF with PFR~\cite{PFR}, Inform~\cite{inform}, and REDRESS~\cite{redress}. 

\begin{table}[!t]
    \caption{Dataset statistics.}
    \label{tab:statistic}
    \centering
    \renewcommand\arraystretch{1}
    \resizebox{\linewidth}{!}{
    \begin{tabular}{c|cccc}
        \toprule
        \textbf{Dataset}   & \textbf{\# Nodes}    & \textbf{\# Edges}    & \textbf{\# Features} & \textbf{\# Classes} \\
        \midrule
            \textbf{Cora}            & 2,708       & 10,556     & 1,433     & 7   \\
            \textbf{CiteSeer}        & 3,327       & 9,104      & 3,703     & 6   \\
            \textbf{PubMed}          & 19,717      & 88,648     & 500       & 3   \\
            \textbf{ACM}             & 16,484      & 71,980     & 8,337     & 9   \\
            \textbf{Coauthor-cs}     & 18,333      & 81,894     & 6,805     & 15  \\
            \textbf{Coauthor-phy}    & 34,493      & 247,962    & 8,415     & 5   \\
        \bottomrule
    \end{tabular}}
\end{table}

\subsection{Experimental Settings}
\subsubsection{Datasets}
We conduct experiments on commonly used six datasets: Cora, Citeseer, Pubmed~\cite{yang2016revisiting}, ACM~\cite{tang2008arnetminer}, coauthor-cs, and coauthor-phy~\cite{shchur2018pitfalls}. For Cora, Citeseer, and Pubmed datasets, we use the public split~\cite{yang2016revisiting}. For the remaining three datasets, we split these datasets with 5\%/10\%/85\% of nodes for training/validation/testing, respectively. Following Dong et al.'s work~\cite{redress}, we employ PCA to reduce the feature dimension of the ACM dataset to 200, in order to save on computational costs. For Coauthor-cs and Coauthor-phy datasets, we utilize the original features. We give a brief overview of these datasets below:

\begin{itemize}
\item \textbf{Cora, Citeseer, and Pubmed} are citation network datasets for text classification. Here, nodes represent documents, and their features correspond to the bag-of-words representation of documents. Edges are the citation links between documents. The goal of these datasets is to classify documents into their respective categories.

\item \textbf{ACM} is a citation network where each node represents a paper, and each edge connecting two nodes represents the citation relationship between two papers. Node features comprise the bag-of-words representation of the abstract of the paper. The objective of the ACM dataset is to predict the category of the paper.

\item \textbf{Coauthor-cs/phy} are two co-authorship networks in which nodes represent authors. An edge is created between two nodes if the authors are co-authors on a paper. Node features are the bag-of-words representation of paper keywords from each author's papers. The task is to predict the more active fields of study for the author.
\end{itemize}

\subsubsection{Evaluation Metrics}
We utilize AUC to evaluate the utility performance. For individual fairness performance evaluation, we follow the settings of the previous work~\cite{redress}, utilizing two commonly used ranking metrics: NDCG@$k$~\cite{ndcg} and ERR@$k$~\cite{err}. These two metrics measure the similarity between the ranking list derived from the oracle matrix $\textbf{S}$ and the outcome similarity matrix $\hat{\textbf{S}}$. For all experiments, $k$ is fixed to 10. SaGIF identifies similar individuals by considering both graph topology and node features. However, to ensure consistency with previous studies and to facilitate a fair comparison, all oracle similarity matrices employed for fairness evaluation (NDCG@10 and ERR@10) are calculated using feature similarity determined by cosine similarity.

\subsubsection{Baselines}
We compare SaGIF with three state-of-the-art individual fairness methods, including PFR~\cite{PFR}, InFoRM~\cite{inform}, and REDRESS~\cite{redress}. A brief overview of these methods is presented as follows:

\begin{itemize}
\item \textbf{PFR} is a pre-process fairness method to encode similar node pairs into a fairness graph, facilitating the learning of pairwise fair data representations. However, PFR is not tailored for the graph-structured data. Thus, we employ PFR to obtain a fair node feature representation with the original features as input and utilize it to train GNNs.

\item \textbf{InFoRM} is the first work to study individual fairness in graph mining and defines the notion of individual fairness in graph mining. This notion leads to a quantitative measure of the potential bias. Motivated by this, InFoRM mitigates biases from input graphs, mining models, and mining results. Here, we employ the individual fairness promotion loss in InFoRM to learn fair GNNs.

\item \textbf{REDRESS} redefines the notion of individual fairness in graphs from a ranking perspective and learns to train fair GNNs by ensuring consistency between ranking lists derived from the oracle similarity matrix and the outcome similarity matrix, respectively.    

\end{itemize}

\subsubsection{Implementation Details}
We conduct all experiments 5 times and reported average results. For all methods, we utilize a 2-layer GCN and a 1-layer SGC as classifiers. All hidden dimensions of GNNs are 16. We utilize the Adam optimizer for all methods including SaGIF. Detailed settings for SaGIF and each baseline method are provided as follows:

\begin{itemize}
\item \textbf{SaGIF}: We set the weight decay and the training epoch to $1 \times 10^{-5}$ and 500, respectively. To find the best performance of SaGIF, the learning rate $lr$ and the dimension of the initial similarity encoding $d_{sim}$ are searched from $\{0.001, 0.005, 0.01, 0.05, 0.1\}$ and $\{8, 16, 32, 64, 128\}$. The balanced hyperparameter $\alpha$ is searched from 0 to 1 except for the hyperparameter sensitivity analysis. We utilize the topology fusion method proposed in Section~\ref{subsec:t_f} to measure among individuals, except for the experiment concerning the impact of similarity metrics. We set $\lambda$ in Eq.~\eqref{eq:prop_mat} to 0.5. When the combination of the main encoder and the classifier constitutes a 2-layer GCN, we employ a 1-layer GCN as the similarity encoder. Conversely, when the combination is a 1-layer SGC, we utilize a 1-layer SGC for the similarity encoder.  
\item \textbf{PFR}: For all datasets, we set the training epoch and weight decay to 3000 and $1\times10^{-5}$. For Cora, Citeseer, and Pubmed datasets, we set the learning rate and $\gamma$ to $1\times10^{-3}$ and $1\times10^{-6}$. For ACM and Coauthor-cs/phy datasets, the learning rate and $\gamma$ are both set to 0.01.
\item \textbf{InFoRM}: For all datasets, we set the training epoch, weight decay, and the learning rate to 3000, $1\times10^{-5}$, and 0.001. For the regularization parameter $\alpha$, search for the best setting, ranging from $1\times10^{-11}$ to $1\times10^{-4}$.
\item \textbf{REDRESS}: For all datasets, we set $\gamma=1$, $k=10$, the learning rate $lr=0.01$, and the weight decay to $5\times10^{-6}$. For Cora, Citeseer, and Pubmed datasets, we set the pre-training and training epochs to $\{300, 30\}$ and $\{0, 500\}$, respectively, for the GCN and SGC backbones. For ACM and Coauthor-cs/phy datasets, we follow the specific hyperparameter settings as per their source code.
\end{itemize}

Moreover, all experiments are conducted on one NVIDIA GeForce RTX 3090 Ti GPU with 24GB memory. All models are implemented with PyTorch and PyTorch-Geometric.

\subsection{Comparison Study}
\label{subsec:comparison}
\begin{table*}[!ht]
\caption{Comparison of SaGIF using GCN backbone with baseline methods on six datasets. In each row, the best result is marked in \textbf{bold}, while the runner-up result is marked with an \underline{underline}.
}
\centering
\label{tab:comparison_gcn}
\renewcommand\arraystretch{0.9}
\begin{tabular}{c|c|cccc|c}
\toprule
\textbf{Datasets}                      & \textbf{Metrics} & \textbf{Vanilla} & \textbf{PFR}          & \textbf{InFoRM}       & \textbf{REDRESS}      & \textbf{SaGIF (Ours)} \\
\midrule
\multirow{3}{*}{\textbf{Cora}}         & AUC     & 95.62 ± 0.13     & 91.25 ± 0.24          & {\ul 95.85 ± 0.19}    & 95.19 ± 0.20          & \textbf{95.89 ± 0.38} \\
                                       & NDCG@10 & 58.57 ± 0.22     & 57.27 ± 0.36          & 58.80 ± 0.08          & \textbf{60.00 ± 0.31} & {\ul 58.88 ± 0.11}    \\
                                       & ERR@10  & 84.46 ± 0.26     & 84.20 ± 0.13          & 84.39 ± 0.09          & {\ul 84.53 ± 0.34}    & \textbf{84.54 ± 0.22} \\
                                       \midrule
\multirow{3}{*}{\textbf{Citeseer}}     & AUC     & 85.07 ± 0.18     & {\ul 85.34 ± 0.19}    & 84.64 ± 0.41          & 84.88 ± 0.57          & \textbf{85.87 ± 0.67} \\
                                       & NDCG@10 & 66.00 ± 0.20     & 65.74 ± 0.23          & \textbf{66.86 ± 0.20} & 66.11 ± 0.74          & {\ul 66.32 ± 0.32}    \\
                                       & ERR@10  & 86.73 ± 0.19     & 86.65 ± 0.28          & \textbf{87.44 ± 0.18} & 87.03 ± 0.31          & {\ul 87.22 ± 0.33}    \\
                                       \midrule
\multirow{3}{*}{\textbf{Pubmed}}       & AUC     & 90.33 ± 0.14     & 87.70 ± 0.17          & {\ul 90.53 ± 0.11} & 89.28 ± 0.15          &  \textbf{90.64 ± 0.21}    \\
                                       & NDCG@10 & 36.36 ± 0.24     & 35.97 ± 0.38          & 36.62 ± 0.17          & {\ul 37.09 ± 0.75}    & \textbf{37.10 ± 0.35} \\
                                       & ERR@10  & 77.34 ± 0.25     & \textbf{77.90 ± 0.44} & 77.47 ± 0.34          & {\ul 77.86 ± 0.37}    & 77.34 ± 0.27          \\
                                       \midrule
\multirow{3}{*}{\textbf{ACM}}          & AUC     & 92.13 ± 0.53     & 92.19 ± 0.28          & {\ul 92.85 ± 0.40} & 91.85 ± 0.38          & \textbf{92.98 ± 0.18}    \\
                                       & NDCG@10 & 33.31 ± 0.36     & {\ul 33.89 ± 0.35}    & 33.65 ± 0.53          & 33.82 ± 0.32          & \textbf{34.23 ± 0.25} \\
                                       & ERR@10  & 76.51 ± 0.14     & 76.72 ± 0.21    & 76.55 ± 0.13          & {\ul 76.77 ± 0.10} & \textbf{76.81 ± 0.13}          \\
                                       \midrule
\multirow{3}{*}{\textbf{Coauthor-cs}}  & AUC     & 99.13 ± 0.12     & 97.66 ± 0.15          & \textbf{99.45 ± 0.04} & {\ul 99.29 ± 0.04}          & 99.23 ± 0.06    \\
                                       & NDCG@10 & 45.60 ± 0.63     & 44.54 ± 0.36          & 51.43 ± 0.24    & {\ul 53.11 ± 0.80} & \textbf{53.14 ± 0.33}          \\
                                       & ERR@10  & 78.88 ± 0.24     & 77.92 ± 0.33          & {\ul 80.76 ± 0.09}    & \textbf{80.78 ± 0.17} & 80.68 ± 0.15          \\
                                       \midrule
\multirow{3}{*}{\textbf{Coauthor-phy}} & AUC     & {\ul 99.56 ± 0.04}     & 95.45 ± 0.09          & {\ul 99.56 ± 0.04}          & 99.43 ± 0.05          & \textbf{99.58 ± 0.03}                      \\
                                       & NDCG@10 & 32.45 ± 0.61     & 26.50 ± 0.33          & {\ul 34.02 ± 1.21}          & \textbf{37.35 ± 0.69}          & 32.28 ± 0.96                      \\
                                       & ERR@10  & {\ul 73.59 ± 0.29}     & 72.81 ± 0.15          & 73.59 ± 0.36          & \textbf{74.45 ± 0.18}          & 73.42 ± 0.08          \\
                                       \bottomrule
\end{tabular}
\end{table*}

\begin{table*}[!t]
\caption{Comparison of SaGIF using SGC backbone with baseline methods on six datasets. In each row, the best result is marked in \textbf{bold}, while the runner-up result is marked with an \underline{underline}.}
\centering
\label{tab:comparison_sgc}
\renewcommand\arraystretch{0.9}
\begin{tabular}{c|c|cccc|c}
\toprule
\textbf{Datasets}                      & \textbf{Metrics} & \textbf{Vanilla}      & \textbf{PFR}          & \textbf{InFoRM}       & \textbf{REDRESS}     & \textbf{SaGIF (Ours)} \\
\midrule
\multirow{3}{*}{\textbf{Cora}}         & AUC     & {\ul 95.66 ± 0.01} & 90.19 ± 0.07          & {\ul 95.66 ± 0.01} & 94.42 ± 0.03         & \textbf{95.84 ± 0.06}           \\
                                       & NDCG@10 & 67.52 ± 0.09          & 58.71 ± 0.11          & 59.24 ± 0.04          & {\ul 70.35 ± 0.04}   & \textbf{71.68 ± 0.28} \\
                                       & ERR@10  & 83.31 ± 0.11          & 85.11 ± 0.24          & 84.55 ± 0.14          & {\ul 84.56 ± 0.09}   & \textbf{85.96 ± 0.41} \\
                                       \midrule
\multirow{3}{*}{\textbf{Citeseer}}     & AUC     & 85.74 ± 0.03          & 84.80 ± 0.19          & {\ul 86.74 ± 0.02}    & 83.10 ± 0.12         & \textbf{87.00 ± 0.13} \\
                                       & NDCG@10 & 63.80 ± 0.08          & {\ul 67.28 ± 0.11}    & 67.02 ± 0.05          & 65.33 ± 0.17         & \textbf{68.10 ± 0.07} \\
                                       & ERR@10  & 81.17 ± 0.26          & {\ul 87.26 ± 0.15}    & \textbf{87.74 ± 0.13} & 82.21 ± 0.15         & 83.59 ± 0.09          \\
                                       \midrule
\multirow{3}{*}{\textbf{Pubmed}}       & AUC     & 90.06 ± 0.01    & 85.45 ± 0.10          & {\ul 90.57 ± 0.01} & 88.41 ± 0.02         & \textbf{90.60 ± 0.05}          \\
                                       & NDCG@10 & 39.33 ± 0.17          & 36.31 ± 0.14          & 36.70 ± 0.04          & \textbf{42.26 ± 0.11}   & {\ul 40.01 ± 0.45} \\
                                       & ERR@10  & 71.86 ± 0.34          & \textbf{77.13 ± 0.57} & {\ul 76.58 ± 0.19}    & 73.14 ± 0.22         & 72.02 ± 0.17          \\
                                       \midrule
\multirow{3}{*}{\textbf{ACM}}          & AUC     & 90.54 ± 0.43          & 91.83 ± 0.27          & {\ul 92.46 ± 0.31}    & 90.55 ± 0.48         & \textbf{93.26 ± 0.26} \\
                                       & NDCG@10 & 57.15 ± 0.59          & 35.36 ± 0.25          & 34.71 ± 0.14          & {\ul 59.65 ± 0.79}   & \textbf{63.13 ± 0.67} \\
                                       & ERR@10  & 80.61 ± 0.30          & 77.32 ± 0.05          & 76.91 ± 0.09          & {\ul 81.72 ± 0.42}   & \textbf{83.29 ± 0.27} \\
                                       \midrule
\multirow{3}{*}{\textbf{Coauthor-cs}}  & AUC     & 98.86 ± 0.10          & 92.16 ± 0.10          & {\ul 99.22 ± 0.06}    & \textbf{99.4 ± 0.04} & 95.65 ± 2.00          \\
                                       & NDCG@10 & 72.56 ± 0.26          & 43.80 ± 0.18          & 53.08 ± 0.19          & {\ul 76.60 ± 0.40}   & \textbf{79.16 ± 5.14} \\
                                       & ERR@10  & 90.22 ± 0.10          & 79.05 ± 0.14          & 81.36 ± 0.06          & {\ul 91.40 ± 0.12}   & \textbf{92.91 ± 1.94} \\
                                       \midrule
\multirow{3}{*}{\textbf{Coauthor-phy}} & AUC     & 99.30 ± 0.04          & 85.42 ± 0.43          & {\ul 99.46 ± 0.04}          & \textbf{99.53 ± 0.03}         & 98.79 ± 0.52  \\
                                       & NDCG@10 & 48.97 ± 0.14          & 26.55 ± 0.16          & 34.66 ± 0.18          & \textbf{52.85 ± 0.54}         & {\ul 50.15 ± 1.27} \\
                                       & ERR@10  & 77.73 ± 0.25          & 72.30 ± 0.11          & 73.52 ± 0.12          & \textbf{79.15 ± 0.38}         &  {\ul 78.12 ± 0.40} \\
                                       \bottomrule
\end{tabular}
\end{table*}
As shown in Tables~\ref{tab:comparison_gcn} and~\ref{tab:comparison_sgc}, we investigate the effectiveness of SaGIF by comparing its performance with baseline methods in the node classification task. For this purpose, a 2-layer GCN and a 1-layer SGC are employed as downstream task models for all methods. The combination of the main encoder and the classifier of SaGIF is consistent with these two downstream task models. Our observations can be summarized as follows: (1) SaGIF outperforms all baseline methods in terms of both fairness and utility in most cases. Although SaGIF may not yield optimal outcomes in some cases, its performance exhibits only a marginal deviation from the most favorable results. (2) In SaGIF, the similarity encoding not only enhances the level of individual fairness but also improves utility, analogous to the role of positional encoding in augmenting the expressive capabilities of GNNs~\cite{dwivedi2021graph}. This observation implies that similarity information is critical in some scenarios, in line with the previous works on similarity preservation of GNNs~\cite{jin2021node}. (3) The SaGIF model, when implemented with an SGC backbone, demonstrates superior performance compared to its implementation with a GCN backbone. The potential explanation is that SaGIF with an SGC backbone, characterized by fewer non-linearities, is advantageous in preserving similarity. To more comprehensively illustrate the performance of SaGIF, we summarize the average ranking of all methods according to Tables~\ref{tab:comparison_gcn} and~\ref{tab:comparison_sgc}, as shown in Table~\ref{tab:avg_rank}. Broadly speaking, in most cases, SaGIF outperforms the baseline methods in terms of the average ranking for both utility and individual fairness. 

\begin{table*}[!t]
    \caption{Average ranking of all methods in Tables~\ref{tab:comparison_gcn} and~\ref{tab:comparison_sgc}. The best result is marked in \textbf{bold}. A.R.: Average Ranking}
    \centering
    \label{tab:avg_rank}
    \renewcommand\arraystretch{1.0}
    \resizebox{0.8\linewidth}{!}{
    \begin{tabular}{c|ccc|ccc}
        \toprule
        \multirow{2}{*}{\textbf{Methods}} & \multicolumn{3}{c}{\textbf{GCN}}                                                                                                   & \multicolumn{3}{c}{\textbf{SGC}}                                                                                                   \\ 
                                          & \multicolumn{1}{c}{\textbf{A.R.:AUC}} & \multicolumn{1}{c}{\textbf{A.R.:NDCG@10}} & \multicolumn{1}{c}{\textbf{A.R.:ERR@10}} & \multicolumn{1}{c}{\textbf{A.R.:AUC}} & \multicolumn{1}{c}{\textbf{A.R.:NDCG@10}} & \multicolumn{1}{c}{\textbf{A.R.:ERR@10}} \\
                                          \midrule
        \textbf{Vanilla}              & 3.17                                    & 4.00                                           & 3.67                                       & 3.17                                    & 3.33                                        & 4.00                                          \\
\textbf{PFR}                      & 4.17                                    & 4.50                                         & 4.00                                          & 4.50                                     & 4.33                                        & 3.17                                       \\
\textbf{InFoRM}                   & 2.33                                    & 2.67                                        & 2.67                                       & \textbf{2.00}                                       & 4.00                                           & 3.33                                       \\
\textbf{REDRESS}                  & 3.83                                    & 2.00                                           & \textbf{1.83}                                       & 3.17                                    & 2.00                                           & 2.50                                        \\
\textbf{SaGIF}                    & \textbf{1.33}                                    & \textbf{1.83}                                        & 2.50                                        & \textbf{2.00}                                       & \textbf{1.33}                                       & \textbf{2.00}                                         \\
        \bottomrule
    \end{tabular}}
\end{table*}

\subsection{Ablation Study}
We conduct ablation studies to gain insights into the effect of three important components of SaGIF on improving IF. We denote SaGIF without $k$NN graph in similarity encoding initialization, similarity encoding initialization, and the similarity-based loss as ``$SaGIF_{-Sk}$'', ``$SaGIF_{-SI}$'' and ``$SaGIF_{-SL}$'', respectively. $SaGIF_{-Sk}$ replaces the $k$NN graph with the original graph to obtain similarity encoding, while $SaGIF_{-SI}$ replaces initial similarity encoding with a random encoding. We conduct experiments on three datasets, employing a 1-layer SGC as the main and similarity encoder.

\begin{figure}[!t]
    \centering
    \includegraphics[width=\linewidth]{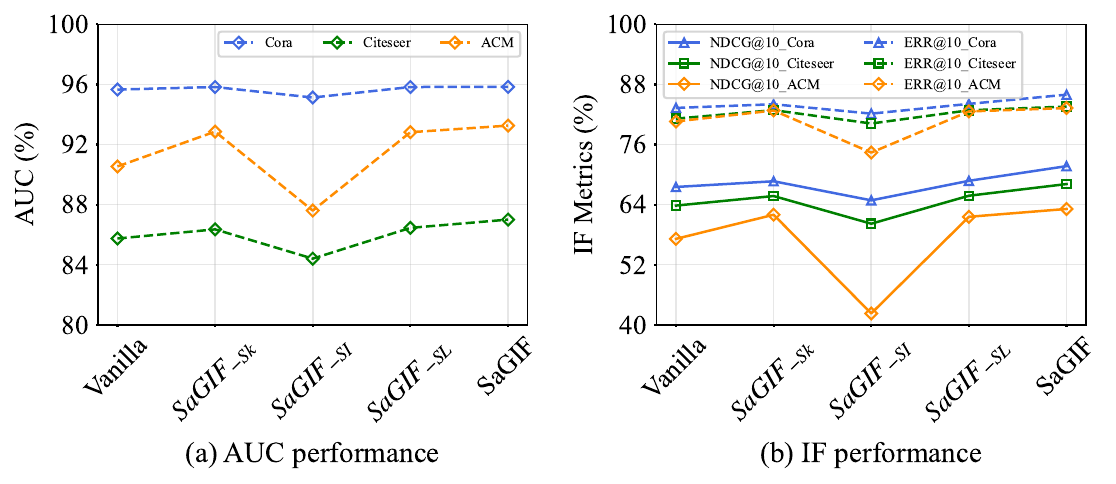}
    \caption{Ablation analysis w.r.t. three components of SaGIF on Cora, Citeseer, and ACM dataset.}
    \label{fig:ablation}
\end{figure}

As shown in Figure~\ref{fig:ablation}, we make the following observations: (1) Removing any one of the three components leads to a decline in SaGIF's performance, which indicates the efficacy of each component. (2) The elimination of the $k$NN graph marginally diminishes SaGIF's performance. This observation implies that the $k$NN graph facilitates preserving similarity information while the inherent similarity information within the original graph still offers a satisfactory initialization. (3) The removal of the similarity encoding initialization has a significant impact on the performance of SaGIF, consistent with prior works~\cite{dwivedi2021graph} integrating GNNs with positional encoding, emphasizing its importance. This phenomenon further shows that similarity encoding effectively captures information aligned with the graph data's inherent structure.

\subsection{Parameters Sensitivity Analysis}
\begin{figure}[!t]
    \centering
    \includegraphics[width=\linewidth]{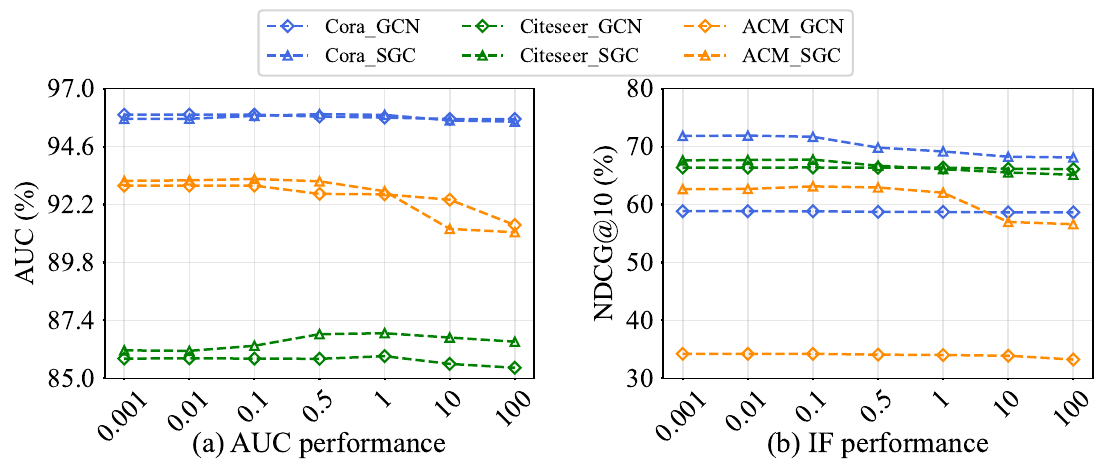}
    \caption{Hyperparameter analysis w.r.t. the balanced hyperparameter $\alpha$ on Cora, Citeseer, and ACM dataset.}
    \label{fig:para_sens}
\end{figure}

We investigate the sensitivity of SaGIF w.r.t. the balanced hyperparameter $\alpha$. We vary the values of $\alpha$ as $\{0.001, 0.01, 0.1, 0.5, 1, 10, 100\}$ on Citeseer and ACM datasets, while maintaining the other hyperparameters as specified in Section~\ref{subsec:comparison}. As shown in Figure~\ref{fig:para_sens}, the overall performance of SaGIF remains stable despite significant variations in $\alpha$. This finding suggests that the similarity encoder consistently preserves similarity even when $\mathcal{L}_{s}$ and its coefficient $\alpha$ are ineffective. This conclusion aligns with the findings presented in Figure~\ref{fig:ablation}, emphasizing the notion that the similarity encoder and its initialization play a more significant role in preserving similarity than the similarity-based loss $\mathcal{L}_{s}$. In addition, we also observe that SaGIF presents a declining trend as $\alpha$ increases, especially on the ACM dataset. This observation indicates that the similarity-based loss merely serves a supplementary function, and acting as the primary loss, it hinders the training of the model.

\subsection{Further Probe}
\label{subsec:probe}
We conduct additional experiments to gain a deeper understanding of SaGIF. Specifically, our additional experiments investigate the impact of $d_{sim}$, fusion types, and the initial similarity encoding on the performance of SaGIF.

\subsubsection{Impact of $d_{sim}$}
We fix other parameters as the same as the comparison study and vary the value of $d_{sim}$ as $\{8,16,32,64,128\}$. We utilize a 1-layer SGC as the main encoder and the similarity encoder. We observe that as the dimensions $d_{sim}$ increase, there is a gradual improvement in fairness performance, while the utility performance remains generally stable in most cases. Meanwhile, the most significant improvement is observed in the ACM dataset. The above observations imply the effectiveness of similarity encoding in preserving similarity, thereby improving individual fairness. It is worth noting that SaGIF exhibits low AUC performance on the ACM dataset when $d_{sim}=8$. In this case, we find that the downstream task loss function $\mathcal{L}_{s}$ fails to converge. Due to the large number of features in the ACM dataset, it is challenging to employ a small $d_{sim}$, e.g., $d_{sim}=8$, to represent similarity information. In this regard, the node similarity representation $\textbf{p}^{l}_{v_i}$ encodes insufficient similarity information that hinders the classifier from making accurate predictions.

\begin{figure}[!t]
    \centering
    \includegraphics[width=\linewidth]{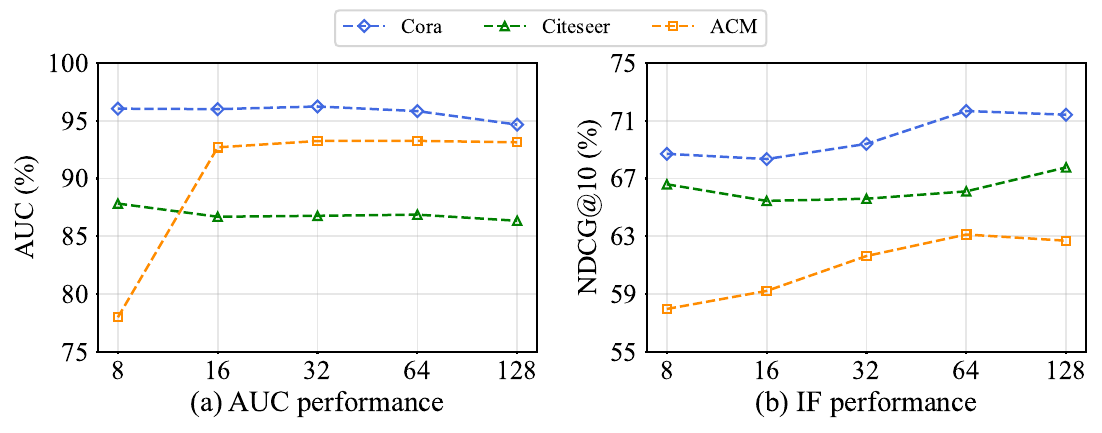}
    \caption{Impact analysis w.r.t. similarity encoding dimensions $d_{sim}$ of SaGIF on Cora, Citeseer, and ACM dataset.}
    \label{fig:dim_impact}
\end{figure}

\subsubsection{Impact of Fusion Types}
\label{subsubsubsec:fusion_types_impact}
As shown in Section~\ref{sec:sim_metric}, there are two methods for constructing the $k$NN graph for the initialization of similarity encoding: feature fusion and topology fusion. Thus, we investigate the impact of these fusion types while maintaining consistent hyperparameter settings as described in Section~\ref{subsec:comparison}. Table~\ref{tab:fusion_impact} presents a comparative analysis of feature fusion and topology fusion, with identical parameter settings for both experimental comparisons, except for the fusion type. It is observed that SaGIF employing topology fusion outperforms its performance when using feature fusion in the majority of cases. However, the overall performance difference between the two fusion approaches is not significant. A possible explanation for these observations is that the Laplacian eigenvector initialization is well-suited for topology fusion.

\begin{table*}[]
\caption{Impact analysis w.r.t. two different fusion types (feature fusion and topology fusion) for constructing $k$NN graph.}
\centering
\label{tab:fusion_impact}
\renewcommand\arraystretch{0.9}
\begin{tabular}{c|c|cc|cc}
\toprule
\multirow{2}{*}{\textbf{Datasets}}     & \multirow{2}{*}{\textbf{Metrics}} & \multicolumn{2}{c}{\textbf{GCN}}                                & \multicolumn{2}{c}{\textbf{SGC}}                              \\
\cmidrule(l){3-4} \cmidrule(l){5-6} 
                                       &                                   & \textbf{SaGIF (feature)} & \textbf{SaGIF (topology)} & \textbf{SaGIF (feature)} & \textbf{SaGIF (topology)} \\
                                       \midrule
\multirow{3}{*}{\textbf{Cora}}         & AUC                      & 95.89 ± 0.38             & \textbf{95.95 ± 0.36}     & \textbf{95.88 ± 0.11}    & 95.84 ± 0.06              \\
                                       & NDCG@10                  & \textbf{58.88 ± 0.11}    & 58.80 ± 0.24              & 71.52 ± 0.14             & \textbf{71.68 ± 0.28}     \\
                                       & ERR@10                   & \textbf{84.54 ± 0.22}    & 84.42 ± 0.20               & 85.71 ± 0.21             & \textbf{85.96 ± 0.41}     \\
                                       \midrule
\multirow{3}{*}{\textbf{Citeseer}}     & AUC                      & 85.87 ± 0.67             & \textbf{85.98 ± 0.63}     & 87.00 ± 0.13             & \textbf{87.05 ± 0.09}     \\
                                       & NDCG@10                  & 66.32 ± 0.32             & \textbf{66.38 ± 0.36}     & \textbf{68.10 ± 0.07}    & 66.91 ± 0.21              \\
                                       & ERR@10                   & \textbf{87.22 ± 0.33}    & 87.16 ± 0.26              & 83.59 ± 0.09             & \textbf{84.02 ± 0.53}     \\
                                       \midrule
\multirow{3}{*}{\textbf{Pubmed}}       & AUC                      & 90.63 ± 0.21             & \textbf{90.64 ± 0.21}     & 90.44 ± 0.19             & \textbf{90.60 ± 0.05}      \\
                                       & NDCG@10                  & 37.06 ± 0.30              & \textbf{37.10 ± 0.35}      & \textbf{40.11 ± 0.73}    & 40.01 ± 0.45              \\
                                       & ERR@10                   & 77.24 ± 0.37             & \textbf{77.34 ± 0.27}     & \textbf{72.16 ± 0.38}    & 72.02 ± 0.17              \\
                                       \midrule
\multirow{3}{*}{\textbf{ACM}}          & AUC                      & 92.98 ± 0.19             & \textbf{92.98 ± 0.18}     & 93.24 ± 0.26             & \textbf{93.26 ± 0.26}     \\
                                       & NDCG@10                  & \textbf{34.23 ± 0.24}    & 34.23 ± 0.25              & 62.68 ± 0.70              & \textbf{63.13 ± 0.67}     \\
                                       & ERR@10                   & 76.80 ± 0.11              & \textbf{76.81 ± 0.13}     & 83.03 ± 0.36             & \textbf{83.29 ± 0.27}     \\
                                       \midrule
\multirow{3}{*}{\textbf{Coauthor-cs}}  & AUC                      & \textbf{99.27 ± 0.06}    & 99.23 ± 0.06              & \textbf{95.65 ± 2.00}    & 95.66 ± 2.01              \\
                                       & NDCG@10                  & 49.32 ± 0.52             & \textbf{53.14 ± 0.33}     & \textbf{79.16 ± 5.14}    & 79.15 ± 5.13              \\
                                       & ERR@10                   & 79.90 ± 0.19              & \textbf{80.68 ± 0.15}     & 92.91 ± 1.94             & \textbf{92.91 ± 1.93}     \\
                                       \midrule
\multirow{3}{*}{\textbf{Coauthor-phy}} & AUC                      & \textbf{99.58 ± 0.03}    & 99.57 ± 0.04              & \textbf{98.79 ± 0.52}    & 98.78 ± 0.51              \\
                                       & NDCG@10                  & \textbf{32.28 ± 0.96}    & 32.25 ± 0.78              & 50.15 ± 1.27             & \textbf{50.16 ± 1.27}     \\
                                       & ERR@10                   & 73.42 ± 0.08             & \textbf{73.50 ± 0.22}     & \textbf{78.12 ± 0.40}    & 78.11 ± 0.41               \\
                                       \bottomrule
\end{tabular}
\end{table*}

\subsubsection{Impact of Initial Similarity Encoding}
In Section~\ref{subsec:init_SE}, SaGIF employs Laplacian eigenvector to initialize similarity encoding. However, such an initialization method is limited by the sign ambiguity~\cite{dwivedi2022benchmarking}. To investigate the effect of initialization methods on SaGIF, we compare the Laplacian eigenvector initialization with Random walk initialization. Specifically, we fix other parameters as the same as the comparison study, except for methods of similarity encoding initialization. Our observations reveal that, in most instances, the performance of Laplacian eigenvector initialization outperforms that of Random walk initialization, particularly when SaGIF employs the SGC backbone. Specifically, Laplacian eigenvector initialization enables the model to learn a unique node similarity representation, while the Random walk initialization offers a similar benefit without necessitating the learning of additional invariances, thereby making it easier to learn. Theoretically, Random walk initialization should yield superior performance; however, the reality contradicts this expectation. A plausible explanation for this discrepancy could be that our employed topology fusion solution is more compatible with Laplacian eigenvector initialization. This conclusion also remains consistent with Section~\ref{subsubsubsec:fusion_types_impact}.

\begin{table*}[!t]
\caption{Impact analysis w.r.t. two different similarity encoding initialization methods (Laplacian eigenvector and Random walk).}
\centering
\label{tab:init_impact}
\renewcommand\arraystretch{0.9}
\begin{tabular}{c|c|cc|cc}
\toprule
\multirow{2}{*}{\textbf{Datasets}}     & \multirow{2}{*}{\textbf{Metrics}} & \multicolumn{2}{c}{GCN}                       & \multicolumn{2}{c}{SGC}                       \\
\cmidrule(l){3-4} \cmidrule(l){5-6}
                                       &                                   & \textbf{SaGIF (lap)}  & \textbf{SaGIF (walk)} & \textbf{SaGIF (lap)}  & \textbf{SaGIF (walk)} \\
                                       \midrule
\multirow{3}{*}{\textbf{Cora}}         & AUC                               & 95.83 ± 0.38          & \textbf{95.89 ± 0.38} & \textbf{95.84 ± 0.06} & 95.74 ± 0.04          \\
                                       & NDCG@10                           & 58.77 ± 0.24          & \textbf{58.88 ± 0.11} & \textbf{71.68 ± 0.28} & 68.05 ± 0.09          \\
                                       & ERR@10                            & \textbf{84.61 ± 0.20} & 84.54 ± 0.22          & \textbf{85.96 ± 0.41} & 83.95 ± 0.18          \\
                                       \midrule
\multirow{3}{*}{\textbf{Citeseer}}     & AUC                               & \textbf{85.87 ± 0.67} & 85.58 ± 0.63          & \textbf{87.00 ± 0.13} & 85.98 ± 0.14          \\
                                       & NDCG@10                           & 66.32 ± 0.32          & \textbf{66.35 ± 0.39} & \textbf{68.10 ± 0.07} & 66.63 ± 0.06          \\
                                       & ERR@10                            & \textbf{87.22 ± 0.33} & 87.17 ± 0.23          & 83.59 ± 0.09          & \textbf{83.83 ± 0.11} \\
                                       \midrule
\multirow{3}{*}{\textbf{Pubmed}}       & AUC                               & \textbf{90.64 ± 0.21} & 90.47 ± 0.2           & \textbf{90.60 ± 0.05}  & 90.52 ± 0.06          \\
                                       & NDCG@10                           & 37.10 ± 0.35           & \textbf{37.23 ± 0.32} & \textbf{40.01 ± 0.45} & 39.51 ± 0.49          \\
                                       & ERR@10                            & \textbf{77.34 ± 0.27} & 77.02 ± 0.42          & 72.02 ± 0.17          & \textbf{72.07 ± 0.32} \\
                                       \midrule
\multirow{3}{*}{\textbf{ACM}}          & AUC                               & \textbf{92.98 ± 0.19} & 92.83 ± 0.14          & \textbf{93.26 ± 0.26} & 91.18 ± 0.38          \\
                                       & NDCG@10                           & \textbf{34.23 ± 0.24} & 34.07 ± 0.22          & \textbf{63.13 ± 0.67} & 55.97 ± 0.92          \\
                                       & ERR@10                            & \textbf{76.80 ± 0.11}  & 76.73 ± 0.12          & \textbf{83.29 ± 0.27} & 80.12 ± 0.51          \\
                                       \midrule
\multirow{3}{*}{\textbf{Coauthor-cs}}  & AUC                               & 99.23 ± 0.06          & \textbf{99.28 ± 0.04} & \textbf{95.65 ± 2.00} & 95.47 ± 2.04          \\
                                       & NDCG@10                           & \textbf{53.14 ± 0.33} & 47.53 ± 1.58          & \textbf{79.16 ± 5.14} & 79.09 ± 4.85          \\
                                       & ERR@10                            & \textbf{80.68 ± 0.15} & 79.56 ± 0.49          & \textbf{92.91 ± 1.94} & 92.87 ± 1.88          \\
                                       \midrule
\multirow{3}{*}{\textbf{Coauthor-phy}} & AUC                               & \textbf{99.58 ± 0.03} & 99.56 ± 0.04          & \textbf{98.79 ± 0.52} & 98.56 ± 0.39          \\
                                       & NDCG@10                           & 32.28 ± 0.96          & \textbf{32.49 ± 0.65} & \textbf{50.15 ± 1.27} & 49.93 ± 1.26          \\
                                       & ERR@10                            & 73.42 ± 0.08          & \textbf{73.58 ± 0.23} & \textbf{78.12 ± 0.40} & 78.00 ± 0.55           \\
                                       \bottomrule
\end{tabular}
\end{table*}

\section{Conclusion}
In this work, we investigate previously unexplored challenges in the individual fairness of GNNs, regarding similar individual identification and underlying reasons behind individual unfairness. To facilitate our investigation, we define similarity consistency as the discrepancy in identifying similar individuals based on graph structure versus node feature. Our preliminary analysis led to a key observation: low similarity consistency harms individual fairness. In light of this observation, we introduce two metrics, topology fusion and feature fusion, to measure individual similarity by considering both node features and graph structure. Subsequently, we propose a similarity-aware GNN for improving individual fairness, namely, SaGIF. The core idea behind SaGIF is to integrate individual similarity via independently learning similarity representations. Experimental results imply that SaGIF outperforms all state-of-the-art IF methods in most cases.

\end{document}